\documentclass{article}

\usepackage{times}
\usepackage{graphicx} 
\usepackage{subfig} 
\usepackage{float}

\usepackage{natbib}

\usepackage{algorithm}
\usepackage{algorithmic}

\usepackage{multirow}

\usepackage{balance}

\usepackage{url}
\usepackage{hyperref}

\usepackage{paralist}

\usepackage{amsmath,amsfonts,amssymb,times,graphicx,natbib,algorithm,algorithmic,hyperref}

\usepackage[accepted]{whi2017}

\usepackage{url}

\usepackage{Definitions}

\def\Zb{\mathbf{Z}}
\def\Sb{\mathbf{S}}

\def\zn{\mathbf{z}_{n}}
\def\Wb{\mathbf{W}}
\def\Ib{\mathbf{I}}
\def\Xb{\mathbf{X}}

\def\s2X{\sigma_x^2}
\def\Yb{\mathbf{Y}}

\def\Bb{\mathbf{B}}

\def\s2B{\sigma_B^2}

\def\Ncal{\mathcal{N}}
\def\Ocal{\mathcal{O}}

\def\Pb{\mathbf{P}}
\def\Lb{\boldsymbol{\lambda}^d}

\newcommand{\xhdr}[1]{\vspace{0.0mm}\noindent{{\bf #1.}}}

\icmltitlerunning{General Latent Feature Model}

\begin{document}

\twocolumn[
\icmltitle{General Latent Feature Modeling for Data Exploration Tasks}

\vspace{-2mm}


\icmlsetsymbol{equal}{*}

\begin{icmlauthorlist}
\icmlauthor{Isabel Valera}{to}
\icmlauthor{Melanie F. Pradier}{ed}
\icmlauthor{Zoubin Ghahramani}{to,goo}
\end{icmlauthorlist}

\icmlaffiliation{to}{University of Cambridge, Cambridge, United Kingdom;}
\icmlaffiliation{goo}{Uber AI Labs, San Francisco, California, USA}
\icmlaffiliation{ed}{Universidad Carlos III de Madrid, Spain}

\icmlcorrespondingauthor{Isabel Valera}{miv24@cam.ac.uk}

\icmlkeywords{interpretability, transparency}

\vskip 0.25in
]

\printAffiliationsAndNotice{}

\begin{abstract} 
\vspace{-1mm}
This paper introduces a general Bayesian nonparametric latent feature model suitable to perform automatic exploratory analysis of heterogeneous datasets, where the attributes describing each object can be either discrete, continuous or mixed variables. 
The proposed model presents several important properties. First, it accounts for heterogeneous data while can be inferred  in linear time with respect to the number of objects and attributes.
Second, its Bayesian nonparametric nature allows us to automatically infer the model complexity from the data, i.e., the number of features necessary to capture the latent structure in the data. 
Third, the latent features in the model are binary-valued variables, easing the interpretability of the obtained latent features in data exploration tasks. 
%

%

\end{abstract} 
\vspace{-5mm}

\section{Introduction}
\vspace{-2mm}
Latent feature models allow us to compact in a few features the immense redundant information present in the observed data, by capturing the statistical dependencies among the different objects and attributes. As a consequence,  they appear as suitable tools to perform data exploratory analysis, i.e, they may help us  to better understand the data \cite{Blanco2012, PsiquiatrasJMLR}. 

There is an extensive literature in latent feature modeling of homogeneous data, where all the attributes describing each object in the database present the same (continuous or discrete) nature. In particular, these works assume that databases contain only either continuous data, usually modeled as Gaussian variables \cite{IBP, Todeschini2013}, or discrete, that can be either modeled by discrete likelihoods \cite{Li2009, PsiquiatrasJMLR, Fran2014} or simply treated as Gaussian variables \cite{Blanco2012, Todeschini2013}.  However, there still exists a lack of works dealing with heterogeneous databases, which in fact are common in real applications.  
As motivating examples, Electronic Health Records from hospitals might contain  lab measurements (often positive real-valued or real-valued data), diagnoses (categorical data) and genomic information (ordinal, count data and categorical data); also, a survey often contains diverse information about the participants such as age (count data), gender (categorical data), salary (positive real data), etc. 
Despite this diversity of data types, the standard approach when dealing with heterogeneous datasets is to treat all the attributes, either continuous or discrete, as  Gaussian variables.

This paper presents a general latent feature model (GLFM) suitable for heterogeneous datasets, where the attributes describing each object can be either discrete, continuous or mixed variables. Specifically, we account for real-valued and positive real-valued as examples of continuous variables, and categorical, ordinal and count data as examples of discrete variables. 
The proposed model extends the essential building block of Bayesian nonparametric latent feature models, the Indian Buffet Process (IBP) by \cite{IBP}, to account for heterogeneous data while maintaining the model complexity of conjugate models.  
Among all the available latent feature models in the literature, we opt for the IBP due to two main reasons.  
First, the nonparametric nature of the IBP allows us to  automatically infer the appropriate model complexity,  i.e., the number of necessary features, from the data.
Second,  the IBP considers binary-valued latent features which has been shown to provide more interpretable results in data exploration than standard real-valued latent feature models~\cite{SuicidasNIPS,PsiquiatrasJMLR}. 
The standard IBP assumes real-valued observations combined with conjugate likelihood models, allowing for fast inference algorithms~\cite{AcceleratedGibbs}. However, we here aim at dealing with heterogeneous databases, such that conjugacy might not be straightforwardly available.
%

\begin{figure*}[ht]
\vspace{-0.25cm}
\subfloat[Graphical model]{\includegraphics[width=0.25\textwidth]{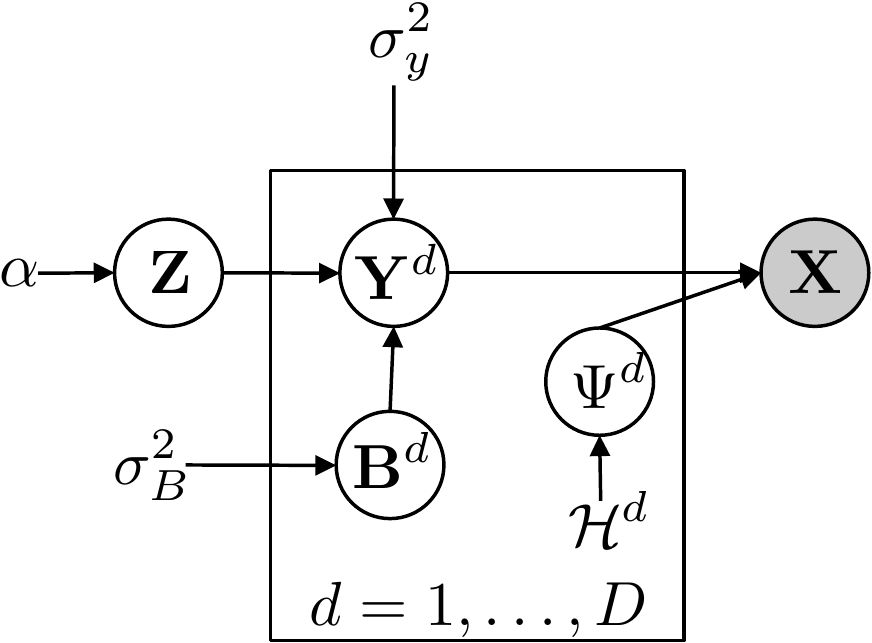}\label{fig:model}} \hspace{10mm}
\subfloat[Example]{\includegraphics[width=0.55\textwidth]{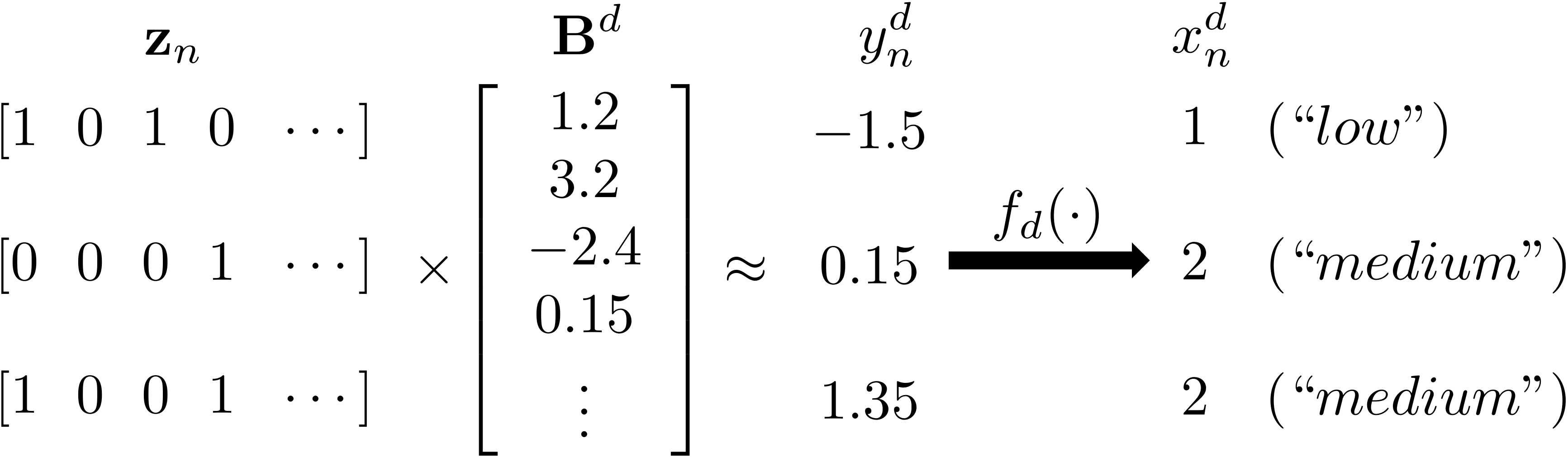}\label{fig:example}}
\centering
\caption{{Illustration of the GLFM.}}
\label{fig:IBPdiscreto}
\vspace{-0.5cm}
\end{figure*}

In order to propose a general observation model for the IBP that accounts for heterogeneous data while keeping the properties of conjugate models, we exploit two key ideas. First, we introduce an auxiliary real-valued variable (also called \emph{pseudo-observation}), such that, conditioned on it, the model behaves as the standard linear-Gaussian IBP in \cite{IBP}.  Second,  we assume that there exists a  function that transforms the pseudo-observation into the actual observation,  mapping the real line into the (discrete or continuous) observation space  of each attribute in the data. 
These two key ideas allow us to derive an efficient inference algorithm based on collapsed Gibbs sampling, which presents linear complexity with the number of objects and attributes in the data. 

Our experiments provide examples of how to use the proposed model for data exploration in real-world datasets. 
Additionally, a software library implementing the GLFM, as well as the necessary scripts to perform automatic data exploration, is publicly available at
\href{https://github.com/ivaleraM/GLFM}{https://github.com/ivaleraM/GLFM} . 

\vspace{-3mm}
\section{General Latent Feature Model}
\vspace{-1mm}
We introduce the GLFM, which is a general Bayesian nonparametric latent feature model suitable for data exploration of heterogeneous datasets, where the attributes describing each object can be either discrete, continuous or mixed variables. 
Specifically, the GLFM accounts for the following data types: \vspace{-2mm}
\begin{compactitem}
\item Continuous variables:
\begin{compactenum}
\item Real-valued, i.e., $x_n^d\in \Re$ 
\item Positive real-valued, i.e., $x_n^d\in \Re_+$. 
\end{compactenum}
\item Discrete variables:
\begin{compactenum}
\item Categorical data, i.e.,  $x_n^d$ takes a value in a finite unordered set, e.g., $x_n^d\in\{$`blue', `red',  `black'$\}$.
\item Ordinal data, i.e.,  $x_n^d$ takes values in a finite ordered set, e.g., $x_n^d\in\{$`never',  `sometimes', `often', `usually', `always'$\}$.
\item Count data, i.e., $x_n^d\in \{0, \ldots, \infty\}$. 
\end{compactenum}
\end{compactitem}

The GLFM builds on the IBP~\cite{IBP}, and therefore, it assumes that each observation $x_n^d$ can be explained by  a potentially infinite-length binary vector $\mathbf{z}_n$ whose elements indicate whether a latent feature is active or not for the $n$-th object; and a (real-valued) weighting vector $\mathbf{B}^d$, whose elements weight the influence of each latent feature in the $d$-th attribute\footnote{For convenience, we here capitalize the vector $\mathbf{B}^d$.}. 
%
Since the product of the latent feature vector and the weighting vector leads to a real-valued variable, it is necessary to map this variable to the desirable output (continuous or discrete) space, for example, the positive real line or the finite ordered set $\{$\textit{low, medium, high}$\}$. Thus, the GLFM assumes the existence of intermediate real-valued auxiliary variables $y_n^d\sim \mathcal{N}(\mathbf{z}^n\mathbf{B}^d, \sigma^2_d)$, called \emph{pseudo-observation}, and a transformation function $f_d(\cdot)$ that maps this variable into the actual observation $x_n^d$, i.e., $x_n^d = f_d(y_n^d+u)$ where $u\sim \mathcal{N}(0, \sigma^2_u)$ is an auxiliary noise with zero mean and small variance $\sigma^2_u$. 
Additionally, the GLFM accounts for a bias term similar to the one in \cite{SuicidasNIPS, PsiquiatrasJMLR}, which corresponds to an extra latent feature that is active for every object in the data and eases the interpretability of the latent features, as shown in next section.  

Figure~\ref{fig:IBPdiscreto} illustrates the GLFM by showing the corresponding graphical model together with an example of the generative model for an ordinal attribute taking values in the ordered set $\{$\textit{low, medium, high}$\}$. 
 The inference of the GLFM is performed using collapsed Gibbs sampling, which presents linear complexity with respect to the number of objects $N$ and the number of attributes $D$. 
Additional details on the model, as well as on the inference algorithm can be found in \cite{ourPaper}.

\vspace{-3mm}
\section{Data Exploration}\label{subsec:data_exploration}
\vspace{-2mm}
%

The main goal of this section is to provide showcase examples about how to include the specific domain knowledge into the proposed GLFM  to  find and analyze the latent structure underlying data in different application domains, i.e., to perform data exploratory analysis.  
In particular, we here show examples of how to select the input data for the GLFM, as well as how to enter these data into the model, in order to obtain interpretable results that can be used  to get a better understanding of the data.
\begin{table*}[t]
\vspace{-2mm}
\centering
\small{\begin{tabular}{ |p{6.5cm}|l|} \hline
Attribute description & Type of variable \\ \hline \hline 
Stage of the cancer & Categorical with 2 categories \\ 
DES treatment level & Ordinal with 3 categories  \\ 
Tumor size in cm$^2$ & Count data  \\ 
Serum Prostatic Acid Phosphatase (PAP)& Positive real-valued \\
Prognosis Status (outcome of the disease) & Categorical with 4 categories \\ 
 \hline
\end{tabular}}
\caption{\textbf{List of considered attributes for the Prostate Cancer dataset.}}
\label{tab:prostate}
\vspace{-5mm}
\end{table*}

\vspace{-3mm}
\subsection{Drug effect in a clinical trial for prostate cancer}
\vspace{-2mm}
Clinical trials are conducted to collect data regarding the safety and efficacy of a new drug before it can be sold in the consumer market, if ever. 
Concretely, the main goal of clinical trials is to prove the efficacy of a new treatment for a disease while ensuring its safety, i.e., check whether its adverse effects 
remain low enough for any dosage level of the drug. 
As an example, the publicly available \textit{Prostate Cancer dataset}\footnote{\href{http://biostat.mc.vanderbilt.edu/wiki/Main/DataSets}{http://biostat.mc.vanderbilt.edu/wiki/Main/DataSets}}  collects data of a clinical trial that aimed at analyzing the effects of  the drug diethylstilbestrol (DES) as a treatment against prostate cancer. 
More in detail, the dataset contains information about 502 patients with prostate cancer in stages\footnote{The stage of a cancer describes the size of a cancer and how far it has grown. Stage 3 means that the cancer is already quite large and may have started to spread into surrounding tissues or local lymph nodes. Stage 4 is more severe, and refers to a cancer that has already spread from where it started to another body organ. This is also called secondary or metastatic cancer. Find more details in~{\href{http://www.cancerresearchuk.org/about-cancer/what-is-cancer/stages-of-cancer}{http://www.cancerresearchuk.org/about-cancer/what-is-cancer/stages-of-cancer}}} 3 and 4, who entered a clinical trial during 1967-1969 and were randomly allocated to different levels of treatment with DES.
The prostate cancer dataset has been used by several studies~\cite{byar1980, Lunn1995} to analyze the survival times of the patients in the clinical trial and the causes behind their death. 
These studies have pointed out that a large dose of the treatment tends to reduce the risk of a cancer death at any time, but also might result in an increased risk of cardiovascular death.

In this section, we apply the proposed GLFM to the {Prostate Cancer dataset} to show that the proposed model can be efficiently used to discover  the statistical dependencies in the data, which in this example corresponds to the effect of the different levels of treatment with DES in the suffering of prostate cancer and cardiovascular diseases. 
%
The  prostate cancer dataset consists of 502 patients and 16 attributes, from which we make use of the five attributes listed in Table~\ref{tab:prostate}. 
The selection of these five attributes allows us not only to reduce the number of local minima in the posterior distribution of the proposed model due to the small sample size of the dataset, but also to focus on capturing the statistical dependencies between the target attributes, i.e, the relationship between the different levels of treatment with DES and the suffering of prostate cancer and cardiovascular diseases.

\begin{figure*}[t]
	\centering
	\subfloat[Type of Cancer]	
	{\includegraphics[width=0.45\textwidth]{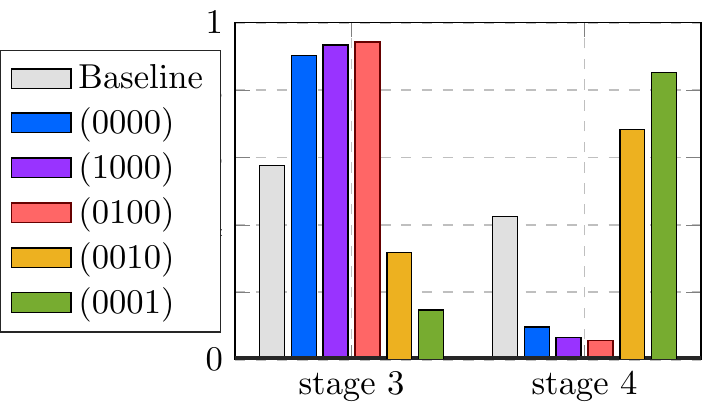}\label{fig:stafe}}\quad\quad
    \subfloat[Drug Level]
    {\includegraphics[width=0.33\textwidth]{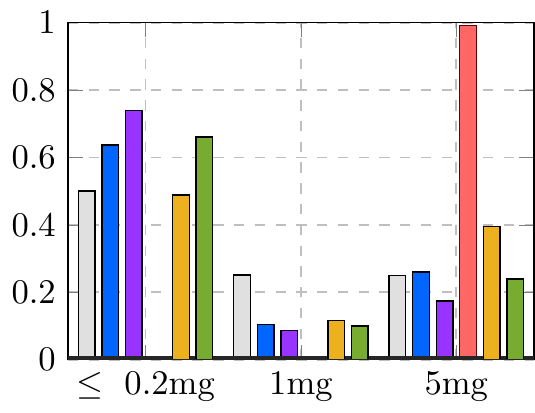}\label{fig:drug}}
    
    \vspace{-0.2cm}
\begin{tabular}{p{0.5\textwidth} p{0.5\textwidth}}
  \vspace{0pt} \subfloat[Size of Primary Tumor (cm$^2$)]
  {\includegraphics[width=0.5\textwidth]{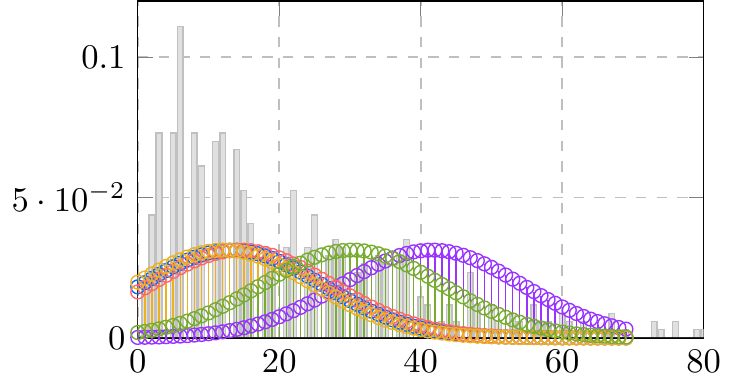}\label{fig:tumor}}  
  &
  \vspace{-1pt} \subfloat[Serum Prostatic Acid Phosphatase]
    {\includegraphics[width=0.45\textwidth]{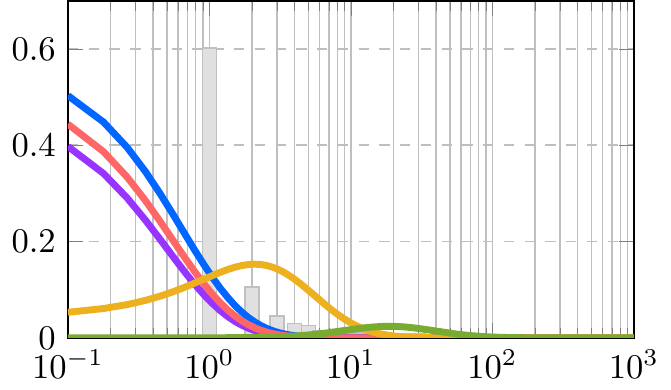}} 
  \label{fig:prosta4}
\end{tabular}

 \vspace{-0.1cm}
\subfloat[Prognosis Status]
  {\includegraphics[width=1\textwidth]{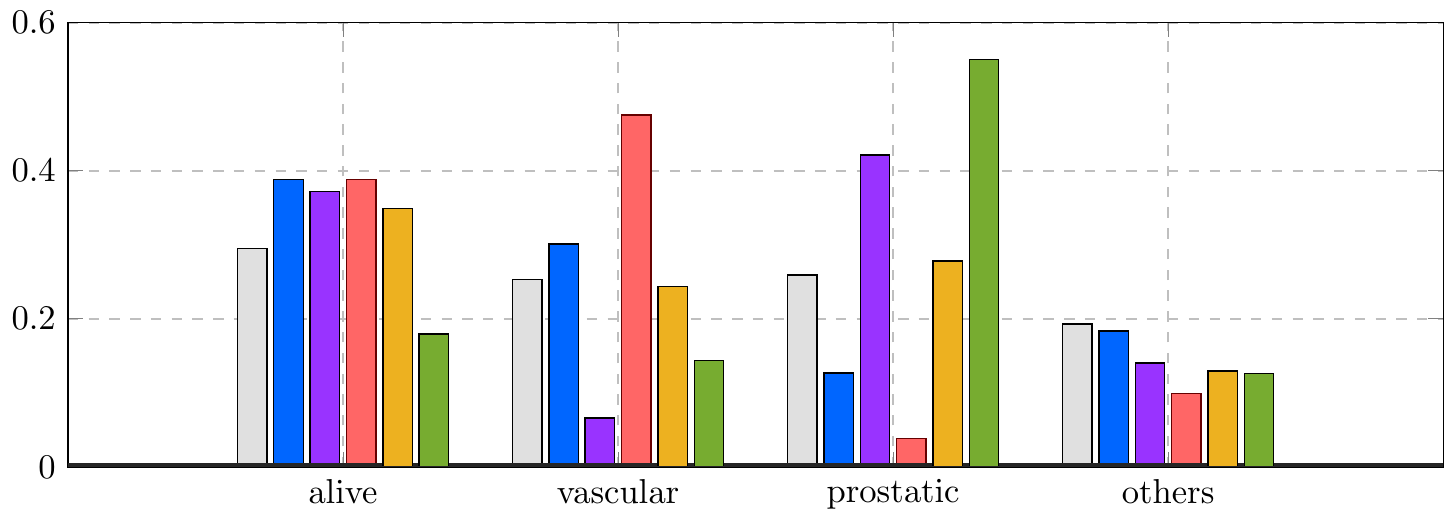}}\label{fig:prosta5}
     \vspace{-0.2cm}
	\caption{\textbf{Data exploration of a prostate cancer clinical trial.} We depict the effect of each latent feature on each attribute. Panels (a)-(d) shows different indicators of the prostate cancer, as well as the dose level of DES. Panel (e) corresponds to Prognosis Status, which indicates whether the patient either is alive or dies from one of the following three causes: vascular disease, prostatic cancer, or other reason. The baseline refers to the empirical distribution of each attribute in the whole dataset.}\label{fig:prostate}
\end{figure*}

\xhdr{Results}
After running our model, we obtain four latent features. 
 Figure~\ref{fig:prostate} shows the effect of the inferred latent features, as well as the bias term, on each dimension/attribute of the data, where we can distinguish two groups of features.
The first group accounts for patients in stage 3 and includes the bias term and the 2 first latent features.
 Within this group,  the bias term -- or equivalently pattern (0000) -- and the first feature -- or equivalently pattern (1000) -- account for patients in stage 3 with a low average level of treatment with DES (refer to Figure~\ref{fig:drug}).
 However, while the bias term models patients with low probability ($\sim15\%$) of prostate cancer death, the first feature accounts for patients with higher probability ($\sim40\%$) of prostate cancer death, which can be explained by a larger tumor size (refer to Figure~\ref{fig:tumor}). 
The second feature -- or equivalently pattern (0100) --captures patients who exclusively received a high dosage (5~mg) of the drug  (refer to Figure~\ref{fig:drug}). 
These patients present a small tumor size and the lowest probability of prostatic cancer death, suggesting a positive effect of the drug as treatment for the cancer. However, they also present a significant increase in the  probability  of dying from a vascular disease ($\sim50\%$), indicating a potential adverse-effect of the drug that increases the risk of suffering from cardio-vascular diseases. Such observation is in agreement with previous studies~\cite{byar1980, Lunn1995}. 

The second group of features corresponds to the activation patterns (0010) and (0001), and accounts for patients in stage 4 with, respectively, mild and severe conditions.
 In particular, the third feature  corresponds to patients with small tumor size, but intermediate values for the PAP biomarker, suggesting a certain spread degree of the tumor compared to the features in the first group, but not as severe as for patients with pattern (0001). Indeed, pattern (0001) models those patients in stage 4 with relatively high tumor size and the highest PAP values--it is thus not surprising that those patients present in turn the highest probability (above 50\%) of prostatic death.
%
 
 %


\vspace{-3mm}
\subsection{Impact of Social Background on Mental Disorders}
\vspace{-2mm}
Several studies have analyzed the impact of social background in the development of mental disorders~\cite{Weissman1993, Weich1998}. 
%
%
%
Other studies have focused on finding and analyzing the co-occurring (comorbidity) pattern among the 20 most common psychiatric disorders ~\cite{Blanco2012, PsiquiatrasJMLR}. 
These studies found that the 20 most common disorders can be divided into three groups:  i) externalizing disorders, which include substance use disorders (alcohol abuse and dependence, drug abuse and dependence and nicotine dependence); ii)  internalizing disorders, which include mood and anxiety disorders (major depressive disorder (MDD), bipolar disorder and dysthymia, panic disorder, social anxiety disorder (SAD), specific phobia and generalized anxiety disorder (GAD), and pathological gambling (PG));  and iii) personality disorders (avoidant, dependent, obsessive-compulsive (OC), paranoid, schizoid, histrionic and antisocial personality disorders (PDs)). 
%
%
However, up to our knowledge, there is a lack of work on the impact of social background in the suffering of comorbid disorders.

In this section, we aim at extending the analysis in~\cite{PsiquiatrasJMLR} to account for  the influence of the social background of subjects (such as age, gender, etc.) in the probability of a subject suffering from  comorbid disorders. 
To this end, in addition to the diagnoses of the above 20 psychiatric disorders, we  also make use of the information provided by the NESARC, 
which includes a set of questions on the social background of participants. Specifically,  in addition to the diagnoses of the most common 20 psychiatric disorders described above, we include  the sex of the participants as input data to the proposed model. 
We model the gender information of the participants in the NESARC as a categorical variable with two categories: $\{$`male', `female'$\}$.  The percentage of males in the NESARC is approximately $ 43\%$.  Note also that the diagnoses of the 20 psychiatric disorders correspond to categorical variables with two possible categories, e.g., a patient suffering or not from a disorder.

\begin{figure*}[]
\vspace*{-5pt}
\centering
\begin{tabular}{c}
\subfloat[Probability of suffering from each disorder]
{\includegraphics[width=1\textwidth]{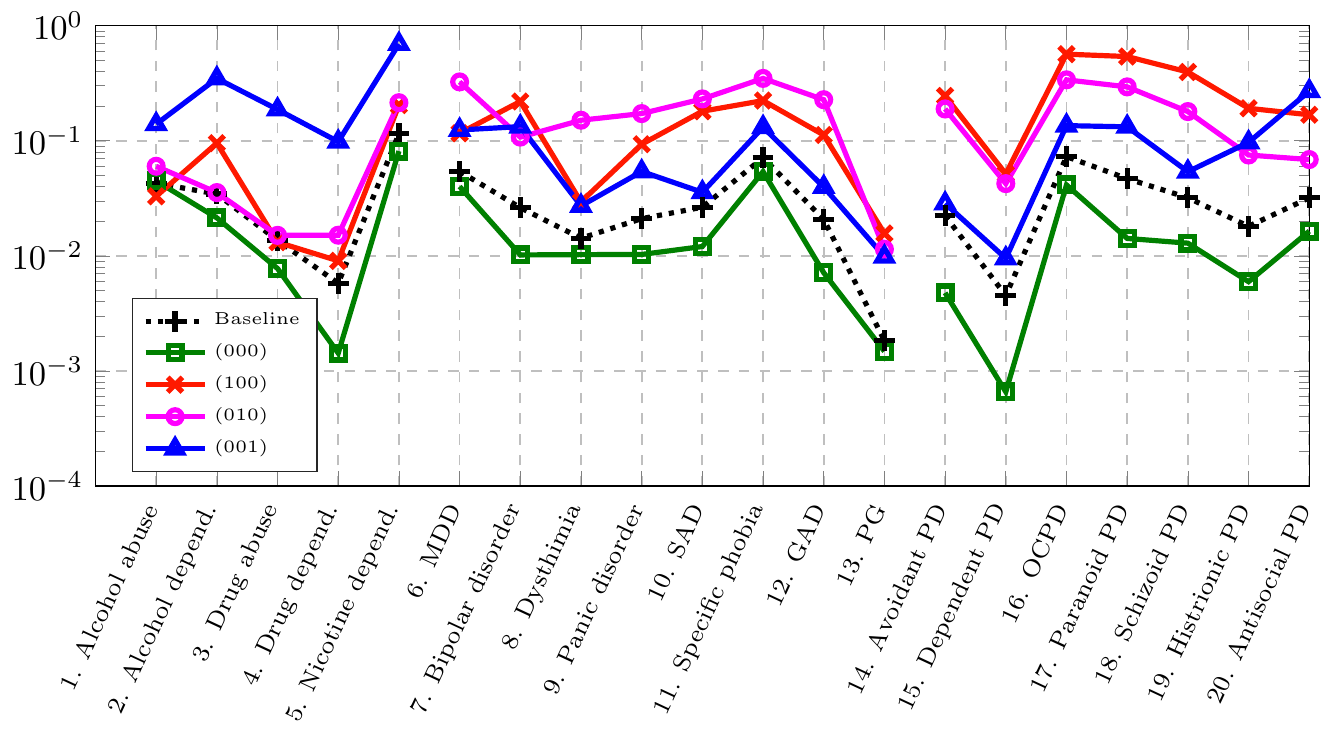}\label{fig:20Q+_5}}\\ 
\subfloat[Gender distribution]
{\includegraphics[width=1\textwidth]{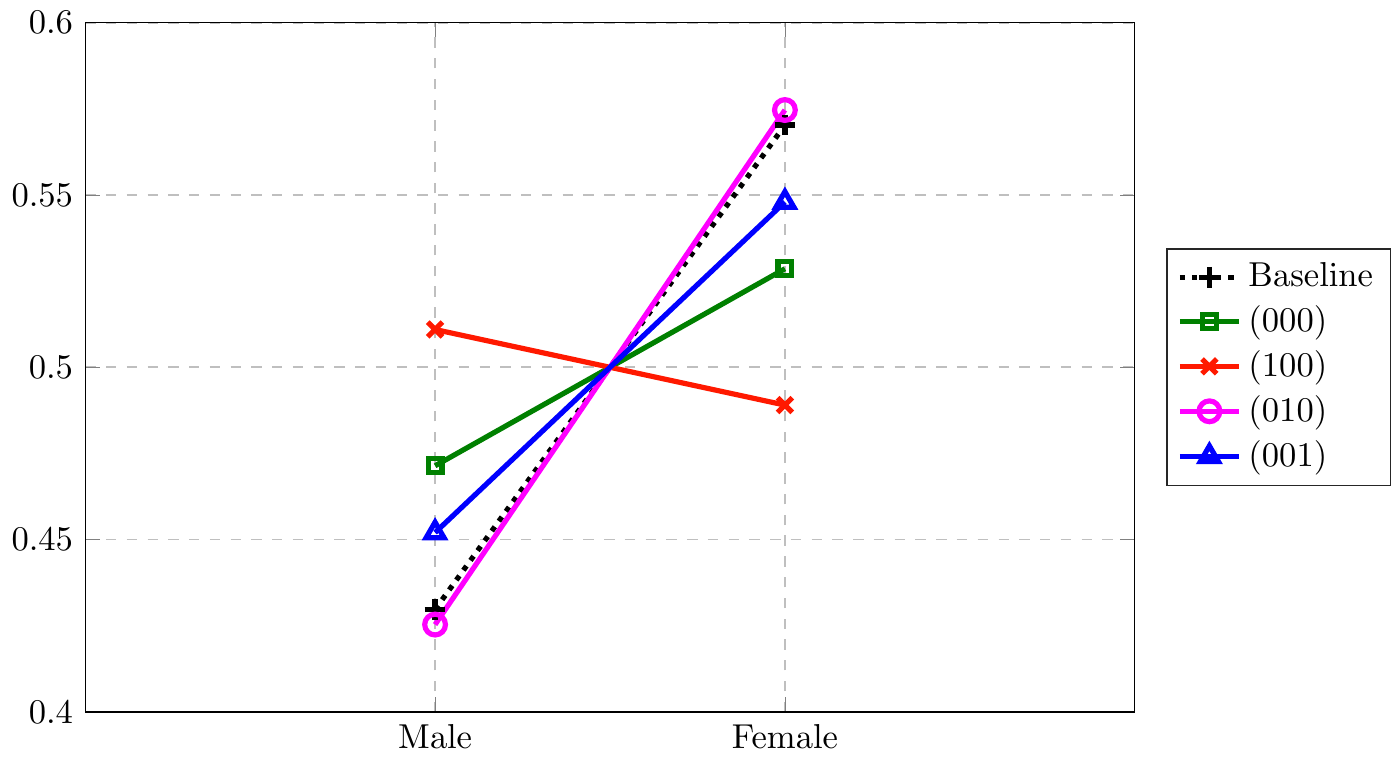} \label{fig:21Q_5}} 
\end{tabular}
\caption{\textbf{Feature effects including gender in the analysis.} (a) Probabilities of suffering from the $20$ considered disorders and (b) probability of being male and female for the latent feature vectors $\zn$ shown in the legend and for the baseline.}\label{fig:gender}
\end{figure*}

\xhdr{Results}
%
After running our inference algorithm with the diagnoses of the $20$ disorders and the gender of subjects as input data, we obtain three latent features. 
%
%
Figure~\ref{fig:20Q+_5} shows the probability of meeting each diagnostic criteria for the latent feature vectors $\zn$ listed in the legend and in the database (baseline). Note that the obtained latent features are similar to the ones in \cite{PsiquiatrasJMLR}, i.e., feature 1 (pattern $(100)$) mainly models the seven personality disorders (PDs), feature 2 (pattern $(010)$) models  alcohol and drug abuse disorders and the antisocial PD, while feature 3 (pattern $(001)$) models  anxiety and mood disorders. Additionally,  Figure~\ref{fig:21Q_5} shows the probability of being male and female for the latent feature vectors $\zn$ depicted in the legend and the empirical probability of being male and female in the database (baseline).

In Figure~\ref{fig:21Q_5}, we observe that having no active features (pattern $(000)$), which captures people that do not suffer from any disorder, increases the probability of being male with respect to the baseline probability, and therefore, it indicates that females tend to suffer in a higher extent from psychiatric disorders. 
Additionally, we observe that  feature 1 (pattern $(100)$) increases the probability of being male, while feature 3 (pattern $(001)$)  increases the probability of being female. Hence, from the analysis of Figure~\ref{fig:21Q_5}, we can conclude that, while women suffer more frequently from mood and anxiety disorders than men, PDs are more common in men.

\vspace{-3mm}
\section{Conclusions}
\vspace{-2mm}
 In this paper, we have introduced the first available general latent feature model and its code implementation, which will ease researchers from diverse fields to analyze a wide range of heterogeneous, incomplete and noisy datasets in an automatic manner. 
 We have showed the flexibility and applicability of the proposed GLFM by performing data exploratory analysis of diverse real-world datasets. Further results including higher dimensional spaces can be found in~\cite{ourPaper}.



\nocite{langley00}

\vspace{-3mm}
\small
\bibliography{Bib}
\bibliographystyle{icml2017}

\end{document}